\documentclass[10pt,twocolumn,letterpaper]{article}

\usepackage{cvpr}
\usepackage{times}
\usepackage{epsfig}
\usepackage{graphicx}
\usepackage{amsmath}
\usepackage{amssymb}
\usepackage{xcolor}
\usepackage{multirow}

\definecolor{darkgreen}{RGB}{147, 196, 125}
\definecolor{pink}{RGB}{244, 204, 204}

\renewcommand{\paragraph}[1]{

        \vspace{3pt}
	\noindent\textbf{#1}}

\usepackage[pagebackref=true,breaklinks=true,letterpaper=true,colorlinks,bookmarks=false]{hyperref}

\cvprfinalcopy 

\ifcvprfinal\fi
\begin{document}

\title{SMOKE: Single-Stage Monocular 3D Object Detection via Keypoint Estimation}
\author{Zechen Liu$^1$ ~~\quad Zizhang Wu$^1$  ~~\quad Roland T\'oth$^2$\\
$^1$ZongMu Tech ~~\quad $^2$TU/e\\
{\tt\small {\{zechen.liu, zizhang.wu\}}@zongmutech.com, R.Toth@tue.nl}
}

\maketitle

\begin{abstract}
    Estimating 3D orientation and translation of objects is essential for infrastructure-less autonomous navigation and driving. In case of monocular vision, successful methods have been mainly based on two ingredients: (i) a network generating 2D region proposals, (ii) a R-CNN structure predicting 3D object pose by utilizing the acquired regions of interest. We argue that the 2D detection network is redundant and introduces non-negligible noise for 3D detection. Hence, we propose a novel 3D object detection method, named SMOKE, in this paper that predicts a 3D bounding box for each detected object by combining a single keypoint estimate with regressed 3D variables. As a second contribution, we propose a multi-step disentangling approach for constructing the 3D bounding box, which significantly improves both training convergence and detection accuracy. In contrast to previous 3D detection techniques, our method does not require complicated pre/post-processing, extra data, and a refinement stage. Despite of its structural simplicity, our proposed SMOKE network outperforms all existing monocular 3D detection methods on the KITTI dataset, giving the best state-of-the-art result on both 3D object detection and Bird's eye view evaluation. The code will be made publicly available.
\end{abstract}

\section{Introduction}

    Vision-based object detection is an essential ingredient of autonomous vehicle perception and infrastructure less robot navigation in general. This type of detection methods are used to perceive the surrounding environment by detecting and classifying object instances into categories and identifying their locations and orientations. Recent developments in 2D object detection \cite{faster_rcnn_2015, ssd_2015, yolo_2016, retinanet_2017, cornernet_2018, centernet_2019} have achieved promising performance on both detection accuracy and speed. In contrast, 3D object detection \cite{mono3d_2016, stereorcnn_2019, voxelnet_2018} has proven to be a more challenging task as it aims to estimate pose and location for each object simultaneously.
    
    Currently, the most successful 3D object detection methods heavily depend on LiDAR point cloud \cite{voxelnet_2018, pointrcnn_2019, std_2019} or LiDAR-Image fusion information \cite{mmf_2019, frustum_2019, MV3D_2017} (features learned from the point cloud are key components of the detection network). However, LiDAR sensors are extremely expensive, have a short service life time and too heavy for autonomous robots. Hence, LiDARs are currently not considered to be economical to support autonomous vehicle operations. Alternatively, cameras are cost-effective, easily mountable and light-weight solutions for 3D object detection with long expected service time. 
    Unlike LiDAR sensors, a single camera in itself can not obtain sufficient spatial information for the whole environment as single RGB images can not supply object location information or dimensional contour in the real world. While binocular vision restores the missing spatial information, in many robotic applications, especially Unmanned Aerial Vehicles (UAVs), it is difficult to realize binocular vision.
    Hence, it is desirable to perform 3D detection on a monocular image even if it is a more difficult and challenging task.
    
    \begin{figure}[t]
    \centering
    \includegraphics[width=\linewidth]{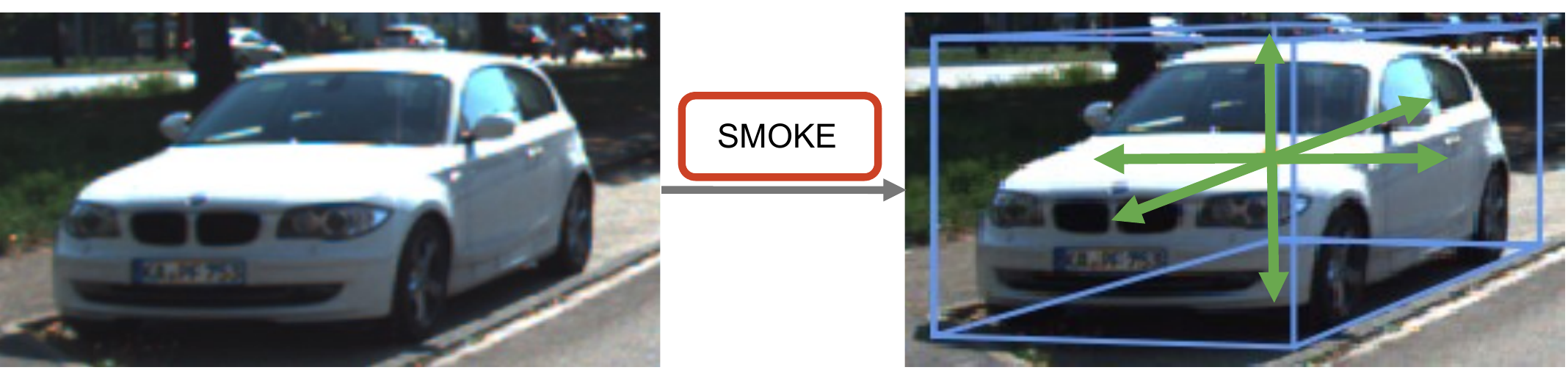}
       \caption{SMOKE directly predicts the 3D projected keypoint and 3D regression parameters on a single image. The whole network is trained end-to-end in a single stage.}
    \label{fig:overview}\vspace{-6mm}
    \end{figure}
    
    \begin{figure*}[t]
    \centering
    \vspace{-4em}
    \includegraphics[width=0.95\linewidth]{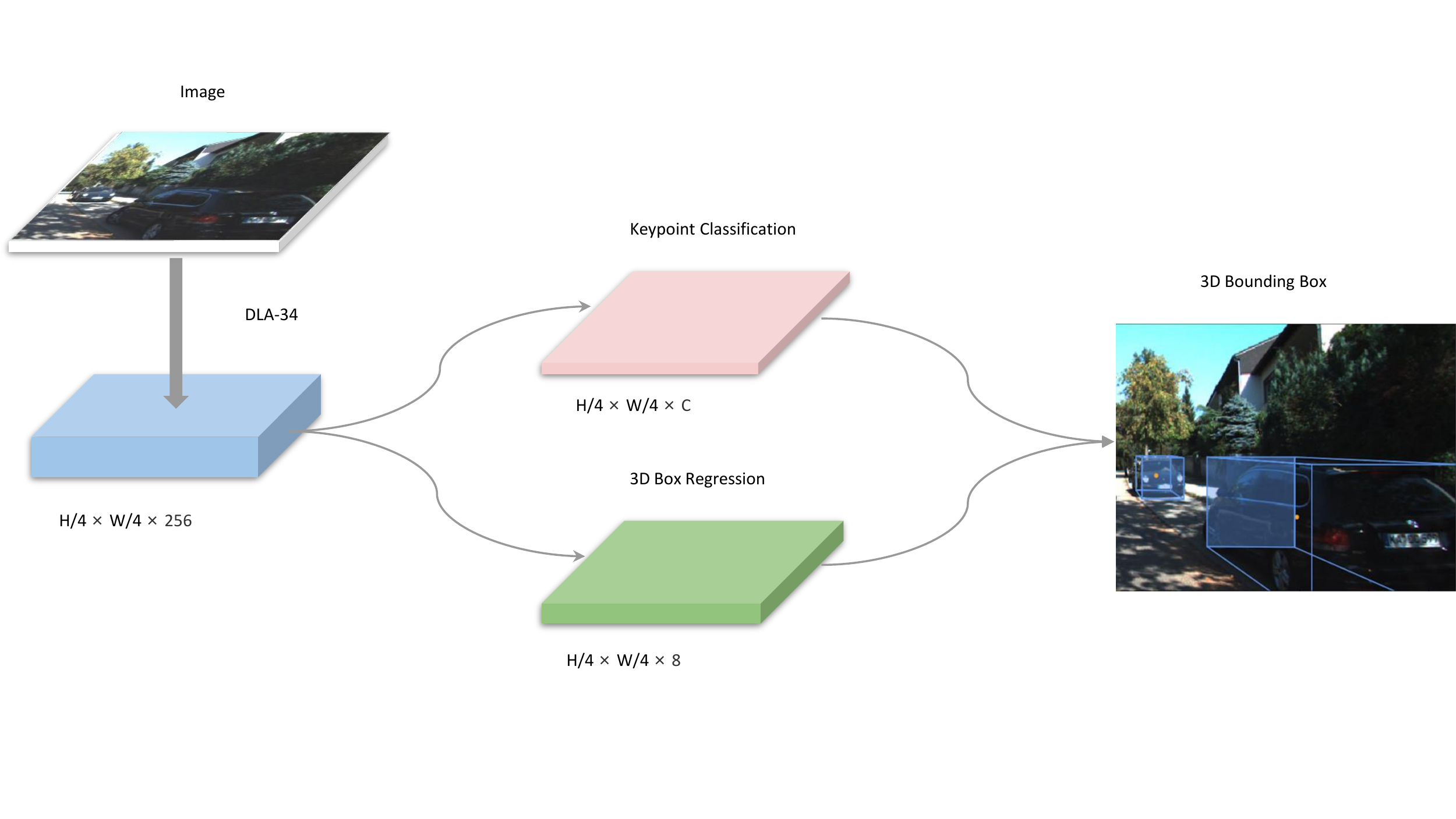}
    \vspace{-4em}
    \caption{\textbf{Network Structure of SMOKE.} We leverage DLA-34 \cite{dla_2018} to extract features from images. The size of the feature map is 1:4 due to  downsampling by 4 of the original image. Two separate branches are attached to the feature map to perform keypoint classification (\textcolor{pink}{pink}) and 3D box regression (\textcolor{darkgreen}{green}) jointly. The 3D bounding box is obtained by combining information from two branches.}
    \label{net_structure} \vspace{-6mm}
    \end{figure*}
    
    Previous state-of-the-art monocular 3D object detection algorithms \cite{monogr2019, m3drpn_2019, am3d_2019} heavily depend on region-based convolutional neural networks (R-CNN) or region proposal network (RPN) structures \cite{faster_rcnn_2015, retinanet_2017, mask_rcnn_2017}. Based on the learned high number of 2D proposals, these approaches attach an additional network branch to either learn 3D information or to generate a pseudo point cloud and feed it into point-cloud-detection network. The resulting multi-stage complex process introduces persistent noise from 2D detection, which significantly increases the difficulty for the network to learn 3D geometry. To enhance performance, geometry reasoning \cite{monogr2019}, synthetic data \cite{roi10d_2019} and post 3D-2D processing \cite{m3drpn_2019} have also been used to improve 3D object detection on single image. By the knowledge of the authors, no reliable monocular 3D detection method has been introduced so far to learn 3D information directly from the image plane avoiding the performance decrease that is inevitable with multi-stage methods.

    In this paper, we propose an innovative single-stage 3D object detection method that pairs each object with a single keypoint. We argue and later show that a 2D detection , which introduces nonnegligible noise in 3D parameter estimation, is redundant to perform 3D object detection. Furthermore, 2D information can be naturally obtained if the 3D variables and camera intrinsic matrix are already known. Consequently, our designed network eliminates the 2D detection branch and estimates the projected 3D points on the image plane instead. A 3D parameter regression branch is added in parallel. This design results in a simple network structure with two estimation threads. Rather than regressing variables in a separate method by using multiple loss functions, we transform these variables together with projected keypoint to 8 corner representation of 3D boxes and regress them with a unified loss function. As in most single-stage 2D object detection algorithms, our 3D detection approach only contains one classification and regression branch. Benefiting from the simple structure, the network exhibits improved accuracy in learning 3D variables, has better convergence and less overall computational needs.
    
    Second contribution of our work is a multi-step disentanglement approach for 3D bounding box regression. Since all the geometry information is grouped into one parameter, it is difficult for the network to learn each variable accurately in a unified way. Our proposed method isolates the contribution of each parameter in both the 3D bounding box encoding phase and the regression loss function, which significantly helps to train the whole network effectively.
    
    Our contribution is summarized as follows:
    \begin{itemize}
        \item We propose a one-stage monocular 3D object detection with a simple architecture that can precisely learn 3D geometry in an end-to-end fashion.
        \item We provide a multi-step disentanglement approach to improve the convergence of 3D parameters and detection accuracy.
        \item The resulting method outperforms all existing state-of-the-art monocular 3D object detection algorithms on the challenging KITTI dataset at the submission date November 12, 2019.
    \end{itemize}

\section{Related Work} \label{sec:1}

In this section, we provide an in-depth overview of the state-of-the-art of 3D object detection based on the used sensor inputs. We first discuss LiDAR based and LiDAR-image fusion methods. After that, stereo image based methods are overviewed. Finally, we summarize approaches that only depend on single RGB images.
    
    \paragraph{LiDAR/Fusion based methods:} 
    LiDAR-based 3D object detection methods achieve high detection precision by processing sparse point clouds into various representations. Some existing methods, e.g., \cite{velofcn_2016, PIXOR_2018}, project point clouds into 2D Bird's eye view and equip standard 2D detection networks to perform object classification and 3D box regression. Others methods, like \cite{voxelnet_2018, pp_2019, 3dfcn_2017, second_2018}, represent point clouds in voxel grid and then leverage 2D/3D CNNs to generate proposals. LiDAR-image fusion methods \cite{mmf_2019, frustum_2019, MV3D_2017} learn relevant features from both the point clouds and the images together. These features are then combined and fed into a joint network trained for detection and classification.  

    \paragraph{Stereo images based methods:} 
    The early work 3DOP \cite{3dop_2015} generates 3D proposals by exploring many handcrafted features such as stereo reconstruction, depth features, and object size priors. TLNet \cite{tinet_2019} introduces a triangulation based learning network to pair detected regions of interests between left and right images. Stereo R-CNN \cite{stereorcnn_2019} creates 2D proposals simultaneously on stereo images. Then, the methods utilize keypoint prediction to generate a coarse 3D bounding box per region. A 3D box alignment w.r.t. stereo images is finally used on the object instance to improve the detection accuracy. Pseudo-LiDAR methods, e.g., \cite{pl_2019}, generate a ``fake'' point cloud and then feed these features into a point cloud based 3D detection network.

    \paragraph{Monocular image based methods:} 
    3D object detection based on a single perspective image has been extensively studied and it is considered to be a challenging task. A common approach is to apply an additional 3D network branch to regress orientation and translation of object instances, see \cite{mono3d_2016, deep3dbox_2017, mf3d_2018, fqnet_2019, GS3D_2019, monogr2019, roi10d_2019, monodis_2019}. Mono3D \cite{mono3d_2016} generates 3D anchors by using massive amount of features via semantic segmentation, object contour, and location priors. These features are then evaluated via an energy function to accommodate learning of relative information. Deep3DBox \cite{deep3dbox_2017} introduces bins based discretization for the estimation of local orientation for each object and 2D-3D bounding box constrain relationships to obtain the full 3D pose. MonoGRNet \cite{monogr2019} subdivides the 3D object localization task into four tasks that estimate instance depth, 3D location of objects, and local corners respectively. These components are then stacked together to refine the 3D box in a global context. The network is trained in a stage-wise fashion and then trained end-to-end to obtain the final result. Some methods, like \cite{subcnn_2017, deepmanta_2017, 3drcnn_2018}, rely on features detected in a 2D object box and leverage external data to pair information from 2D to 3D. DeepMANTA \cite{deepmanta_2017} proposes a coarse-to-fine process to generate accurate 2D object proposals, which proposals are then used to match a 3D CAD model from an external annotated dataset. 3D-RCNN \cite{3drcnn_2018} also uses 3D models to pair the outputs from a 2D detection network. They then recover the 3D instance shape and pose by deploying a render-and-compare loss. Other approaches, like \cite{am3d_2019, monopl_2019, monopsr_2019}, generate hand-crafted features by transforming region of interest on images to other representations. AM3D transforms 2D imagery to a 3D point cloud plane by combining it with a depth map. A PointNet \cite{pointnet_2017} is then used to estimate 3D dimensions, locations and orientations. The only one-stage method M3D-RPN \cite{m3drpn_2019} proposes a standalone network to generate 2D and 3D object proposals simultaneously. They further leverage a depth-aware network and post 3D-2D optimization technique to improve precision. OFTNet \cite{oft_2019} maps the 2D feature map to bird-eye view by leveraging orthographic feature transform and regress each 3D variable independently. Consequently, none of the above methods can estimate 3D information accurately without generating 2D proposals.

    
\section{Detection Problem} \label{Sec:detect}
    We formulate the monocular 3D object detection problem as follows: given a single RGB image $\textit{I} \in \mathbb{R}^{W \times H \times 3}$, with $W$ the width and $H$ the height of the image, find for each present object its category label $C$ and its 3D bounding box $B$, where the latter is parameterized by 7 variables $(h, w, l, x, y, z, \theta)$. Here, $(h, w, l)$ represent the height, weight, and length of each object in meters, and $(x, y, z)$ is the coordinates (in meters) of the object center in the camera coordinate frame. Variable $\theta$ is the yaw orientation of the corresponding cubic box. The roll and pitch angles are set to zero by following the KITTI \cite{kitti} annotation. Additionally, we take the mild assumption that the camera intrinsic matrix \textit{K} is known for both training and inference.
    \begin{figure}[t]
        \centering
        \includegraphics[width=\linewidth]{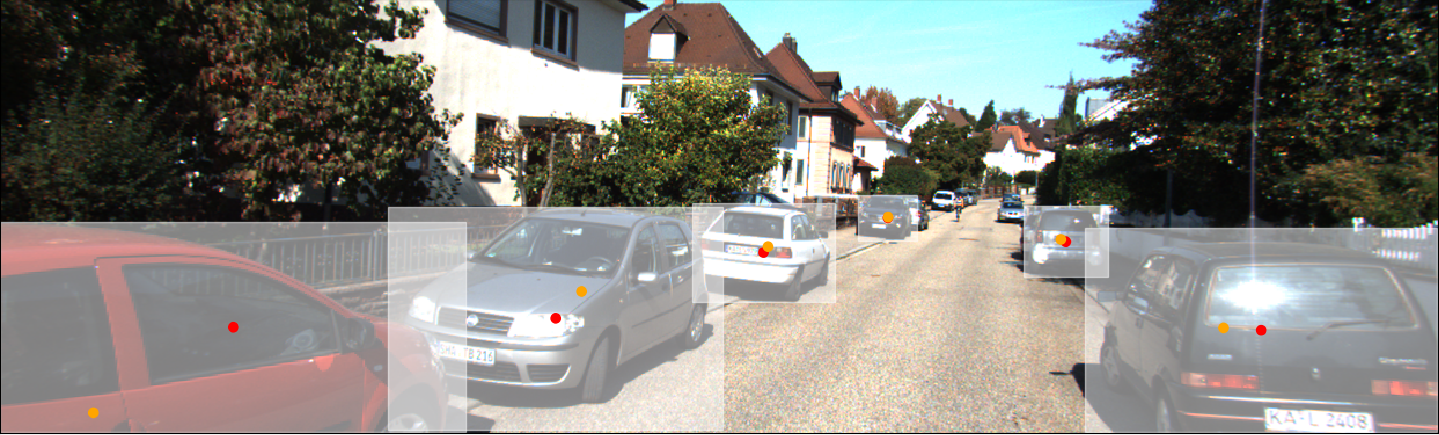}
        \caption{Visualization of difference between 2D center points (\textcolor{red}{red}) and 3D projected points (\textcolor{orange}{orange}). Best viewed in color.} \vspace{-4mm}
        \label{2d_3d_points}
    \end{figure}
    
\section{SMOKE Approach} \label{sec:SMOKE}
    In this section, we describe the SMOKE network that directly estimates 3D bounding boxes for detected object instances from monocular imagery. In contrast to previous techniques that leverage 2D proposals to predict a 3D bounding box, our method can detect 3D information with a simple single stage. The proposed method can be divided into three parts: (i) backbone, (ii) 3D detection, (iii) loss function. First, we briefly discuss the backbone for feature extraction, followed by the introduction of the 3D detection network consisting of two separated branches. Finally, we discuss the loss function design and the multi-step disentanglement to compute the regression loss. The overview of the network structure is depicted in Fig. \ref{net_structure}.
    
\subsection{Backbone}
    We use a hierarchical layer fusion network DLA-34 \cite{dla_2018} as the backbone to extract features since it can aggregate information across different layers. Following the same structure as in \cite{centernet_2019}, all the hierarchical aggregation connections are replaced by a Deformable Convolution Network (DCN) \cite{dcn_2019}. The output feature map is downsampled 4 times with respect to the original image. Compared with the original implementation, we replace all BatchNorm (BN) \cite{bn_2015} operation with GroupNorm (GN) \cite{gn_2018} since it has been proven to be less sensitive to batch size and more robust to training noise.  We also use this technique in the two prediction branches, which will be discussed in Sec.~\ref{3d_net}. This adjustment not only improves detection accuracy, but it also reduces considerably the training time. In Sec.~\ref{abla_sty}, we provide performance comparison of BN and GN to demonstrate these properties.
    
    \begin{figure}
    \centering
    \vspace{-2em}
    \includegraphics[width=0.9\linewidth]{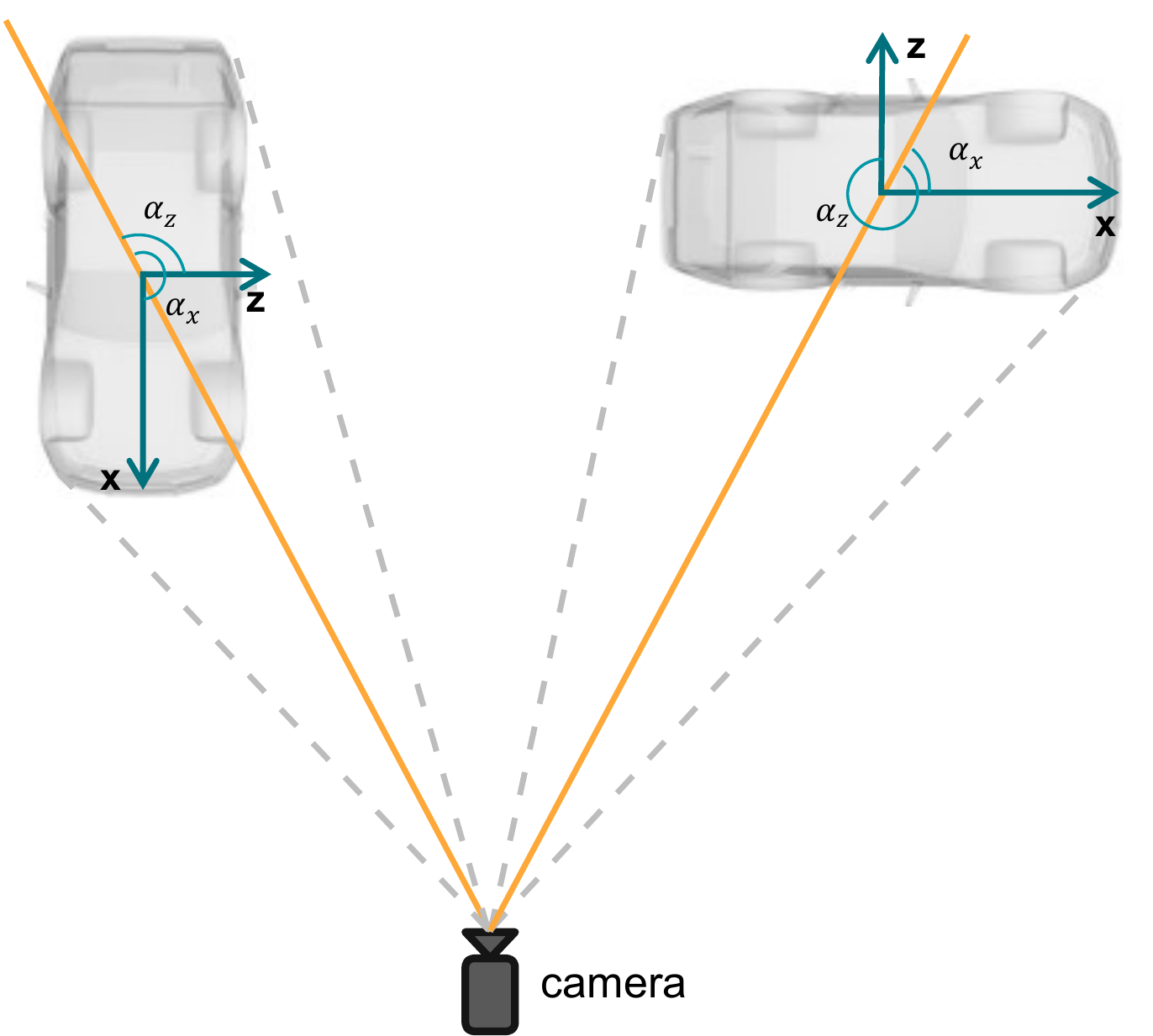}
    \caption{Relation of the observation angle $\alpha_x$ and $\alpha_z$. $\alpha_x$ is provided in KITTI, while $\alpha_z$ is the value we choose to regress.} \vspace{-3mm}
    \label{rotation_difference}
    \end{figure}
    
\subsection{3D Detection Network}
\label{3d_net}
    \paragraph{Keypoint Branch:} 
    We define the keypoint estimation network similar to \cite{centernet_2019} such that each object is represented by one specific keypoint. Instead of identifying the center of a 2D bounding box, the key point is defined as the projected 3D center of the object on the image plane. The comparison between 2D center points and 3D projected points is visualized in Fig.~\ref{2d_3d_points}. The projected keypoints allow to fully recover 3D location for each object with camera parameters. Let $\begin{bmatrix} x \!&\!  y \!&\! z  \end{bmatrix}^{\top}$ represent the 3D center of each object in the camera frame. The projection of 3D points to points $\begin{bmatrix} x_\mathrm{c} \!&\! y_\mathrm{c} \end{bmatrix}^\top$ on the image plane can be obtained with the camera intrinsic matrix \textit{K} in a homogeneous form:

    \begin{equation}
        \begin{bmatrix} 
        z \cdot x_\mathrm{c} \\
        z \cdot y_\mathrm{c} \\
        z
        \end{bmatrix} = 
        K_{3 \times 3}     \begin{bmatrix} 
                                x \\
                                y \\
                                z
                           \end{bmatrix}.
   \label{projection}
    \end{equation}

    For each ground truth keypoint, its corresponding downsampled location on the feature map is computed and distributed using a Gaussian Kernel following \cite{centernet_2019}. The standard deviation is allocated based on the 3D bounding boxes of the ground truth projected to the image plane. Each 3D box on the image is represented by 8 2D points $\begin{bmatrix} {x}_{\mathrm{b},1 \sim 8} \!&\! {y}_{\mathrm{b},1 \sim 8} \end{bmatrix}^\top$ and the standard deviation is computed by the smallest 2D box with $\{ {x}_\mathrm{b}^\mathrm{min} ,{y}_\mathrm{b}^\mathrm{min}, {x}_\mathrm{b}^\mathrm{max}, {y}_\mathrm{b}^\mathrm{max} \}$ that encircles the 3D box.
    
    \paragraph{Regression Branch:} 
    Our regression head predicts the essential variables to construct 3D bounding box for each keypoint on the heatmap. Similar to other monocular 3D detection framework \cite{roi10d_2019, monodis_2019}, the 3D information is encoded as an 8-tuple  $\tau = \begin{bmatrix}\delta_{z} \ \ \delta_{x_\mathrm{c}} \ \ \delta_{y_\mathrm{c}}\ \ \delta_{h} \ \ \delta_{w} \ \ \delta_{l}\ \ \sin \alpha\ \ \cos \alpha \end{bmatrix}^\top$. Here $\delta_{z}$ denotes the depth offset, $\delta_{x_\mathrm{c}}, \delta_{y_\mathrm{c}}$ is the discretization offset due to downsampling, $\delta_h, \delta_w, \delta_l$ denotes the residual dimensions, $\sin (\alpha), \cos (\alpha)$ is the vectorial representation of the rotational angle $\alpha$. We encode all variables to be learnt in residual representation to reduce the learning interval and ease the training task. The size of feature map for regression results in $S_\mathrm{r} \in \mathbb{R}^{\frac{H}{R} \times \frac{W}{R} \times 8}$. Inspired by the lifting transformation described in \cite{roi10d_2019}, we introduce a similar operation $\mathcal{F}$ that converts projected 3D points to a 3D bounding box $B = \mathcal{F}(\tau) \in \mathbb{R}^{3 \times 8}$. For each object, its depth $z$ can be recovered by pre-defined scale and shift parameters $\sigma_z$ and $\mu_z $ as
    \begin{equation}
        z = \mu_z + \delta_z\sigma_z.
    \end{equation}
    Given the object depth $z$, the location for each object in the camera frame can be recovered by using its discretized projected centroid $\begin{bmatrix} x_\mathrm{c}\ \ y_\mathrm{c} \end{bmatrix}^\top$ on the image plane and the downsampling offset $\begin{bmatrix} \delta_{x_\mathrm{c}}\ \ \delta_{y_\mathrm{c}} \end{bmatrix}^\top$: 
    \begin{equation}
        \begin{bmatrix} 
            x \\
            y \\
            z
       \end{bmatrix} = K_{3\times 3}^{-1}    \begin{bmatrix} 
                                    z \cdot (x_\mathrm{c} + \delta_{x_\mathrm{c}}) \\
                                    z \cdot (y_\mathrm{c} + \delta_{y_\mathrm{c}}) \\
                                    z
                                 \end{bmatrix}.
    \label{inverse_projection}
    \end{equation}
    This operation is the inverse of Eq.~\eqref{projection}.
    In order to retrieve object dimensions $\begin{bmatrix} h\ \ w\ \ l \end{bmatrix}^\top$, we use a pre-calculated category-wise average dimension $\begin{bmatrix} \Bar{h}\ \ \Bar{w}\ \ \Bar{l} \end{bmatrix}^\top$ computed over the whole dataset. Each object dimension can be recovered by using the residual dimension offset $\begin{bmatrix} \delta_h\ \ \delta_w\ \ \delta_l \end{bmatrix}^\top$:

    \begin{equation}
        \begin{bmatrix} 
            h \\
            w \\
            l
       \end{bmatrix} =     \begin{bmatrix} 
                                \Bar{h}\cdot e^{\delta_h} \\
                                \Bar{w}\cdot e^{\delta_w} \\
                                \Bar{l}\cdot e^{\delta_l}
                           \end{bmatrix}.
    \label{dimension}
    \end{equation}
    Inspired by \cite{deep3dbox_2017}, we choose to regress the observation angle $\alpha$ instead of the yaw rotation $\theta$ for each object. We further change the observation angle with respect to the object head $\alpha_x$, instead of the commonly used observation angle value $\alpha_z$, by simply adding $\frac{\pi}{2}$. The difference between these two angles is shown in Fig.~\ref{rotation_difference}. Moreover, each $\alpha$ is encoded as the vector $\begin{bmatrix} \sin (\alpha)\ \ \cos (\alpha) \end{bmatrix}^\top$. The yaw angle $\theta$ can be obtained by utilizing $\alpha_z$ and the object location:
    
    \begin{equation}
        \theta = \alpha_z + \arctan\left(\frac{x}{z} \right).\label{eq:ang}
    \end{equation}
    Finally, we can construct the 8 corners of the 3D bounding box in the camera frame by using the yaw rotation matrix $R_{\theta}$, object dimensions $\begin{bmatrix} h\ \ w\ \ l \end{bmatrix}^\top$ and location  $\begin{bmatrix} x\ \ y\ \ z \end{bmatrix}^\top$:

    \begin{equation}
        B = R_{\theta}     \begin{bmatrix} 
                                \pm h/2 \\
                                \pm w/2 \\
                                \pm l/2
                           \end{bmatrix} + \begin{bmatrix} 
                                                x \\
                                                y \\
                                                z
                                           \end{bmatrix}.
    \end{equation}

\subsection{Loss Function}
    \paragraph{Keypoint Classification Loss:} 
    We employ the penalty-reduced focal loss \cite{cornernet_2018, centernet_2019} in a point-wise manner on the downsampled heatmap. Let $s_{i,j}$ be the predicted score at the heatmap location $(i, j)$ and $y_{i,j}$ be the ground-truth value of each point assigned by Gaussian Kernel. Define $\breve{y}_{i,j}$ and $\breve{s}_{i,j}$ as:
    \begin{equation*}
        \breve{y}_{i,j}\!=\! 
        \begin{cases}
            0& \ \text{if}\ y_{i,j} = 1\\
            y_{i,j}& \ \text{otherwise}
        \end{cases}, \ \ 
        \breve{s}_{i,j}\! =\! 
        \begin{cases}
            s_{i,j}& \  \text{if}\ y_{i,j} = 1\\
            1 - s_{i,j}& \  \text{otherwise}
        \end{cases},
    \end{equation*}
    
    For simplicity, we only consider a single object class here. Then, the classification loss function is constructed as
    \begin{equation}
        L_\mathrm{cls} = -\frac{1}{N} \sum_{i,j=1}^{h,w}
        (1 - \breve{y}_{i,j})^\beta(1 - \breve{s}_{i,j})^\alpha\mathrm{log}(\breve{s}_{i,j}),
        \label{focal_loss}
    \end{equation}
    where $\alpha$ and $\beta$ are tunable hyper-parameters and $N$ is the number of keypoints per image. The term $(1 - y_{i,j})$ corresponds to penalty reduction for points around the groundtruth location.
    
    \paragraph{Regression Loss:} 
    We regress the 8D tuple $\tau$ to construct the 3D bounding box for each object. We also add channel-wise activation to the regressed parameters of dimension and orientation at each feature map location to preserve consistency. The activation functions for the dimension and the orientation are chosen to be the sigmoid function $\sigma$ and the $\ell_2$ norm, respectively:
    \begin{equation*}
        \begin{bmatrix} 
            \delta_h \\
            \delta_w \\
            \delta_l
       \end{bmatrix} = \sigma\left(\begin{bmatrix} 
                                    o_h \\
                                    o_w \\
                                    o_l
                               \end{bmatrix}\right) - \frac{1}{2}, \quad
        \begin{bmatrix} 
            \sin \alpha \\
            \cos \alpha \\
       \end{bmatrix} =  \begin{bmatrix} 
                            o_{\sin}/\sqrt{o_{\sin}^2 + o_{\cos}^2}  \\
                            o_{\cos}/\sqrt{o_{\sin}^2 + o_{\cos}^2} \\
                       \end{bmatrix},   
    \end{equation*}
    Here $o$ stands for the specific output of network. By adopting the keypoint lifting transformation introduced in Sec.~\ref{3d_net}, we define the 3D bounding box regression loss as the $\ell_1$ distance between the predicted transform $\hat{B}$ and the groundtruth $B$:
\begin{equation}
    L_\mathrm{reg} = \frac{\lambda}{N} \lVert \hat{B} - B \rVert_1,
\end{equation}
    where $\lambda$ is a scaling factor. This is used to ensure that neither the classification, nor the regression dominates the other. The disentangling transformation of loss has been proven to be an effective dynamic method to optimize 3D regression loss functions in \cite{monodis_2019}. Following this design, we extend the concept of loss disentanglement into a multi-step form. In Eq.~\eqref{inverse_projection}, we use the projected 3D groundtruth points on the image plane $\begin{bmatrix} x_\mathrm{c} \ \ y_\mathrm{c} \end{bmatrix}^\top$ with the network predicted discretization offset $\begin{bmatrix} \hat{\delta}_{x_c} \ \ \hat{\delta}_{y_c} \end{bmatrix}^\top$ and depth $\hat{z}$ to retrieve  the location $\begin{bmatrix} \hat{x} \ \ \hat{y}\ \ \hat{z} \end{bmatrix}^\top$ of each object. In Eq.~\eqref{eq:ang}, we use the groundtruth location $\begin{bmatrix} x\ \ y\ \ z \end{bmatrix}^\top$ and the predicted observation angle $\hat{\alpha}_z$ to construct the estimated yaw orientation $\hat{\theta}$. The 8 corners representation of the 3D bounding box is also isolated into three different groups following the concept of disentanglement, namely orientation, dimension and location. The final loss function can be represented by:
    \begin{equation}
        L = L_\mathrm{cls} + \sum_{i=1}^3 L_\mathrm{reg}(\hat{B}_i),
    \end{equation}
    where $i$ represents the number of groups we define in the 3D regression branch. The multi-step disentangling transformation divides the contribution of each parameter group to the final loss. In Sec.~\ref{abla_sty}, we show that this method significantly improves detection accuracy.

\subsection{Implementation }
    \begin{table*}[ht]
        \centering
        \begin{tabular}{c||c||c||ccc||ccc}
        \hline
        \multirow{2}{*}{Method} & \multirow{2}{*}{Backbone} & \multirow{2}{*}{Runtime(s)} & \multicolumn{3}{c||}{3D Object Detection}                                   & \multicolumn{3}{c}{Birds' Eye View}                                       \\ \cline{4-9} 
                                &                           &                             & \multicolumn{1}{c}{Easy} & \multicolumn{1}{c}{Moderate} & Hard           & \multicolumn{1}{c}{Easy} & \multicolumn{1}{c}{Moderate} & Hard           \\ \hline
        OFTNet\cite{oft_2019}                & ResNet-18                 & 0.50                        & 1.32                      & 1.61                          & 1.00           & 7.16                      & 5.69                          & 4.61          \\
        GS3D\cite{GS3D_2019}                 & VGG-16                    & 2.00                        & 4.47                      & 2.90                          & 2.47           & 8.47                      & 6.08                          & 4.94           \\
        MonoGR\cite{monogr2019}              & VGG-16                    & 0.06                        & 9.61                      & 5.74                          & 4.25           & 18.19                     & 11.17                         & 8.73          \\
        ROI-10D\cite{roi10d_2019}            & ResNet-34                 & 0.20                        & 4.32                      & 2.02                          & 1.46           & 9.78                      & 4.91                          & 3.74          \\
        MonoDIS\cite{monodis_2019}           & ResNet-34                 & 0.10                        & 10.37                     & 7.94                          & 6.40           & 17.23                     & 13.19                         & 11.12          \\
        M3D-RPN\cite{m3drpn_2019}            & DenseNet-121              & 0.16                        & \textbf{14.76}            & 9.71                          & 7.42           & \textbf{21.02}            & 13.67                         & 10.23          \\ \hline \hline
        Ours                                 & DLA-34                    & \textbf{0.03}               & 14.03                     & \textbf{9.76}                 & \textbf{7.84}  & 20.83                     & \textbf{14.49}                & \textbf{12.75} \\ \hline
        \end{tabular}
        \vspace{2mm}
        \caption{\textbf{Test set performance.} 3D object detection and Birds' eye view performance w.r.t.~the car class on the official KITTI data set using the \textit{test} split. Both metrics are evaluated by $\text{AP}|_{R_{40}}$ at 0.7 IoU threshold.}
        \label{testresult} \vspace{-3mm}
    \end{table*}
    
     \begin{table}[t]
        \begin{tabular}{c||ccc}
        \hline
        \multirow{2}{*}{Method} & \multicolumn{3}{c}{3D Object Detection / Birds' Eye View}          \\ \cline{2-4} 
                                        & Easy          & Moderate      & Hard          \\ \hline
        CenterNet\cite{centernet_2019}  & 0.86 / 3.91   & 1.06 / 4.46   & 0.66 / 3.53   \\
        Mono3D\cite{mono3d_2016}        & 2.53 / 5.22   & 2.31 / 5.19   & 2.31 / 4.13   \\
        OFTNet\cite{oft_2019}           & 4.07 / 11.06  & 3.27 / 8.79   & 3.29 / 8.91   \\
        GS3D\cite{GS3D_2019}            & 11.63 / -     & 10.51 / -     & 10.51 / -     \\
        MonoGR\cite{monogr2019}         & 13.88 / -     & 10.19 / -     & 7.62 / -      \\
        ROI-10D\cite{roi10d_2019}       & 9.61 / 14.50  & 6.63 / 9.91   & 6.29 / 8.73   \\
        MonoDIS\cite{monodis_2019}& 18.05 / 24.26 & 14.98 / 18.43 & 13.42 / 16.95 \\
        M3D-RPN\cite{m3drpn_2019} & 20.40 / 26.86 & 16.48 / 21.15 & 13.34 / 17.14 \\ \hline \hline
        Ours                      & 14.76 / 19.99 & 12.85 / 15.61 & 11.50 / 15.28 \\ \hline
        \end{tabular} \vspace{-1mm}
        \caption{\textbf{Validation set performance.} 3D object detection and Birds' eye view performance w.r.t.~the car class on the official KITTI data set using the \textit{val} split. Both metrics are evaluated by $\text{AP}|_{R_{11}}$ at 0.7 IoU threshold.}
        \label{valresult}
        \vspace{-3mm}
    \end{table}

    In this section, we discuss the implementation of our proposed methodology in detail together with selection of the hyperparemeters.

    \paragraph{Preprocessing:} 
    We avoid applying any complicated preprocessing method on the dataset. Instead, we only eliminate objects whose 3D projected center point on the image plane is out of the image range. Note that the total number of projected center points outside the image boundary for the \textit{car} instance is 1582. This accounts for only the 
    5.5\% of the entire set of 28742 labeled cars

    \paragraph{Data Augmentation:} 
    Data augmentation techniques we used are random horizontal flip, random scale and shift. The scale ratio is set to 9 steps from 0.6 to 1.4, and the shift ratio is set to 5 steps from -0.2 to 0.2. Note that the scale and shift augmentation methods are only used for heatmap classification since the 3D information becomes inconsistent with data augmentation.

    \paragraph{Hyperparameter Choice:} 
    In the backbone, the group number for GroupNorm is set to 32. For channels less than 32, it is set to be 16. For Eq.~\eqref{focal_loss}, we set $\alpha = 2$ and $\beta = 4$ in all experiments. Based on \cite{monodis_2019}, the reference car size and depth statistics we use are $\begin{bmatrix} \Bar{h} \ \ \Bar{w} \ \ \Bar{l} \end{bmatrix}^\top = [1.63 \ \ 1.53\ \ 3.88]^\top$ and $\begin{bmatrix} \mu_z \ \ \sigma_z \end{bmatrix}^\top = [28.01 \ \ 16.32]^\top$ (measured in meters).

    \paragraph{Training:} 
    Our optimization schedule is easy and straightforward. We use the original image resolution and pad it to 1280 $\times$ 384. We train the network with a batch size of 32 on 4 Geforce TITAN X GPUs for 60 epochs. The learning rate is set at $2.5 \times 10^{-4}$ and drops at 25 and 40 epochs by a factor of 10. During testing, we use the top 100 detected 3D projected points and filter it with a threshold of 0.25. No data augmentation method and NMS are used in the test procedure. Our implementation platform is Pytorch 1.1, CUDA 10.0, and CUDNN 7.5.

\section{Performance Evaluation} \label{sec:Eval}

    We evaluate the performance of our proposed framework on the challenging KITTI dataset. The KITTI dataset is a broadly used open-source dataset to evaluate visual algorithms on a driving scene considered representative for autonomous driving. It contains 7481 images for training and 7518 images for testing. The test metric is divided into \textit{easy}, \textit{moderate} and \textit{hard} cases based on the height of the 2D bounding box of object instances, occlusion and truncation level. Frequently, the training set is split into 3712 training examples and 3769 validation examples as mentioned in \cite{mono3d_2016}. For the 3D detection task of our proposed method, the 3D Object Detection and Bird's Eye View benchmarks are available for evaluation. 
    
\subsection{Detection on KITTI}
    \begin{table}[t]
    \centering
    \vspace{2mm}
    \begin{tabular}{c||ccc}
    \hline
    \multirow{2}{*}{Method} & \multicolumn{3}{c}{2D Object Detection} \\ \cline{2-4} 
                              & Easy         & Moderate    & Hard         \\ \hline
    Mono3D\cite{mono3d_2016}  & 94.52        & 89.37       & 79.15        \\
    OFTNet\cite{oft_2019}     & -            & -           & -            \\
    GS3D\cite{GS3D_2019}      & 86.23        & 76.35       & 62.67        \\
    MonoGR\cite{monogr2019}   & 88.65        & 77.94       & 63.31       \\
    ROI-10D\cite{roi10d_2019} & 76.56        & 70.16       & 61.15       \\
    MonoDIS\cite{monodis_2019}& 94.61        & 89.15       & 78.37       \\
    M3D-RPN\cite{m3drpn_2019} & 89.04        & 85.08       & 69.26       \\ \hline \hline 
    Ours                      & 92.88        & 86.95       & 77.04       \\ \hline
    \end{tabular}
    \vspace{3mm}
    \caption{\textbf{2D detection.} $\text{AP}|_{R_{40}}$ performance w.r.t.~the  car  class  on  the  official KITTI data set using the \textit{test} split.}
    \label{2d_detection} \vspace{-3mm}
    \end{table}
    \paragraph{3D Object Detection Performance:} 
    The 3D detection results of our proposed method on the split sets \textit{test} and \textit{val} are compared with the state-of-the-art single image-based methods in Tabs.~\ref{testresult} and \ref{valresult}. We principally focus on the \emph{car} class since it has been at the focus of previous cooperative studies. For both tasks, the \emph{average precision} (AP) with \emph{Intersection over Union} (IoU) larger than 0.7 is used as the metric for evaluation. Note that as pointed out by \cite{monodis_2019}, the official KITTI evaluation has been using 40 recall points instead of 11 recall points to measure the AP value since October 8, 2019. However, previous methods only report accuracy at 11 points on the \textit{val} set. For fair comparison, we report the average precision on 40 points $\text{AP}|_{R_{40}}$ on the \textit{test} set and $\text{AP}|_{R_{11}}$ on the \textit{val} set.
    
    Results on the \textit{test} split, shown in Tab.~\ref{testresult}, show that SMOKE outperforms all existing monocular methods on both 3D object detection and Bird's eye view evaluation metrics. We achieve improvement in the moderate and hard sets and comparable results on the easy set in the 3D object detection task. For Bird's eye view detection, we also achieve notable improvement on the moderate and hard sets. Compared with other methods that increase image size for better performance, our approach uses relatively low-resolution input and still achieves competitive results on the hard set in 3D detection. Next to these, SMOKE shows a significant improvement on detection speed. Without the time-consuming region proposal process and by the benefits of single-stage structure, our proposed method only needs 30ms to run on a TITAN XP. Note that we only compare our method with methods that directly learn features from images. Approaches based on hand-crafted features \cite{monopl_2019, am3d_2019} are not listed in the table. However, with respect to the \textit{val} set of KITTI, the performance degrades as reported in Tab.~\ref{valresult}. We argue that this is due to a lack of training objects. A similar problem has been reported in \cite{centernet_2019}.

\begin{figure}
    \centering
        \vspace{-2em}
        \includegraphics[width=\linewidth]{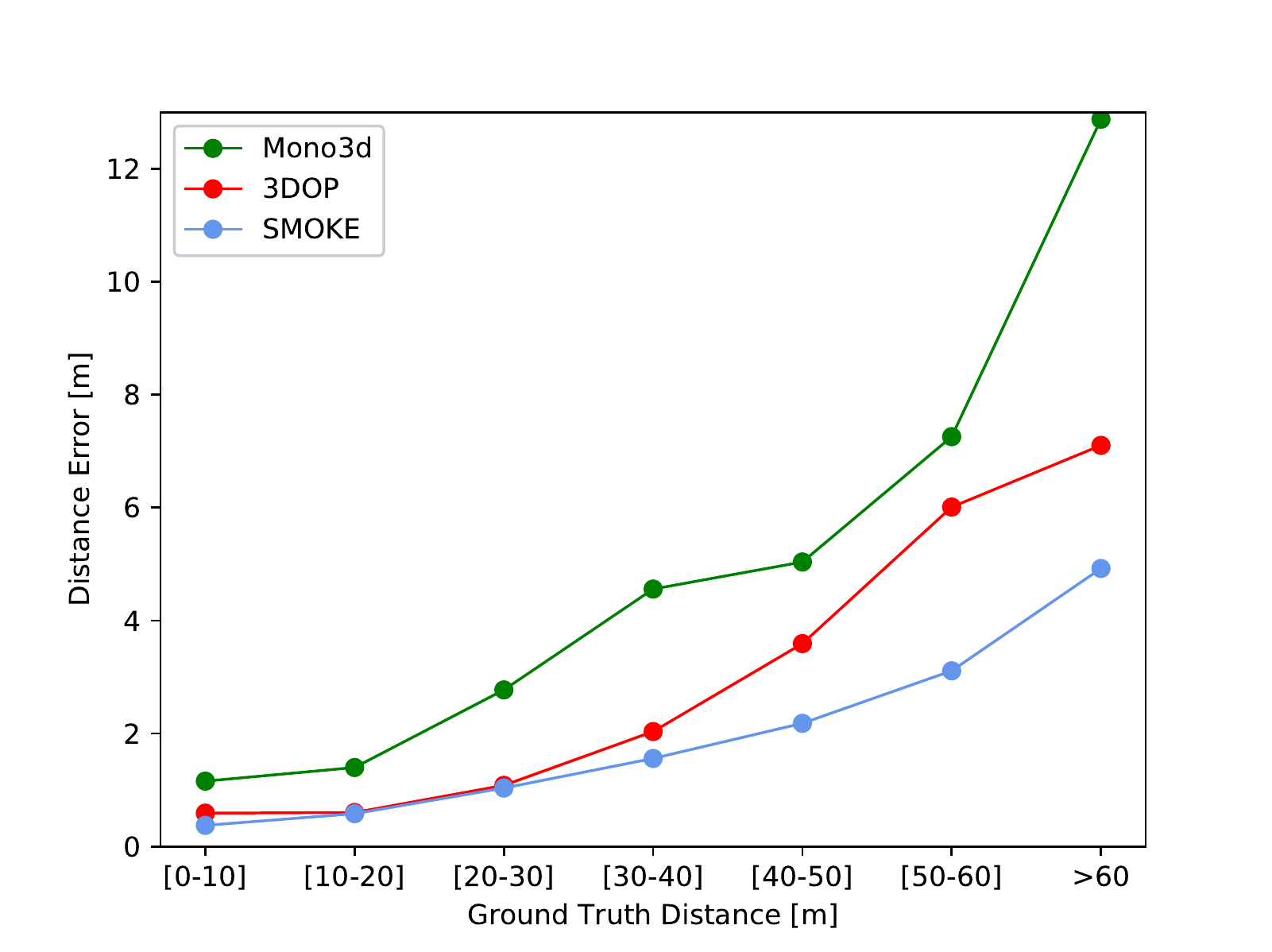}
       \caption{Average depth estimation error visualized in intervals of 10 meters. Best viewed in color.}
        \label{depth_error} \vspace{-3mm}
    \end{figure}

    Estimation of object location in a monocular image is difficult since the incompleteness of spatial information. We evaluate the depth estimation of SMOKE using two different distance measures. In Fig.~\ref{depth_error}, the achieved depth error is displayed in intervals of 10 meters. The error is computed if the 2D bounding box of a detection with any of the ground truth objects has an IoU larger than 0.7. As shown in the figure, the depth estimation error increases as the distance grows. This phenomenon has been observed in many monocular image-based detection algorithms since small objects have large distance distribution. We compare our method with two other methods Mono3D \cite{mono3d_2016} and 3DOP \cite{3dop_2015} on the same \textit{val} set. The curve indicates that our proposed SMOKE method outperforms both methods largely on depth error. Especially at distances larger than 40m, our method achieves more robust and accurate depth estimation.
    
    \paragraph{2D Object Detection:} 
    The 2D detection performance on the official KITTI \textit{test} set is depicted in Tab.~\ref{2d_detection}. Although the 2D bounding box is not directly regressed in the SMOKE network, we observe that our method achieves comparative results on the 2D object detection task. The 2D detection box is obtained as the smallest rectangle that encircles the projected 3D bounding box on the image plane. Unlike other approaches following a 2D$\rightarrow$3D structure, our proposed method reverse this process in a 3D$\rightarrow$2D fashion and outperforms many of the existing methods. This clearly shows that 3D object detection provides more abundant information than 2D detection, hence 2D proposals are redundant and not needed for 3D detection. Furthermore, our proposed method does not use extra data, complicated networks and high-resolution input compared to other methods.

\subsection{Ablation Study}

\label{abla_sty}
    \begin{table}[t]
    \centering
        \begin{tabular}{c||ccc}
        \hline
        \multirow{2}{*}{Option} & \multicolumn{3}{c}{3D Object Detection / Birds' Eye View}          \\ \cline{2-4} 
                        & Easy                                & Moderate                         & Hard        \\ \hline
        BN              & 8.20 / 17.85                        & 8.27 / 15.46                     & 6.50 / 15.21   \\
        GN              & \textbf{10.60} / \textbf{18.06}     & \textbf{8.33} / \textbf{16.07}   & \textbf{6.98} / \textbf{15.39}   \\ \hline
    
        \end{tabular}\vspace{2mm}
        \caption{\textbf{Normalization Strategy.} GN perfoms better than BN on all difficulty sets and in both evaluation metrics.}
        \label{normal_strat} \vspace{-3mm}
    \end{table}

    \begin{table}[t]
    \centering
        \begin{tabular}{c||ccc}
        \hline
        \multirow{2}{*}{Option} & \multicolumn{3}{c}{3D Object Detection / Birds' Eye View}          \\ \cline{2-4} 
                        & Easy                                & Moderate                         & Hard        \\ \hline
        Smooth $\ell_1$       & 10.60 / 18.06                       & 8.33 / 16.07                     & 6.98 / 15.39   \\
        $\ell_1$              & 11.03 / \textbf{20.90}              & 10.53 / 15.95                    & 9.14 / \textbf{15.57}   \\ 
        Dis.~$\ell_1$          & \textbf{14.76} / 19.99              & \textbf{12.85} / \textbf{16.07}  & \textbf{11.50} / 15.39   \\ \hline    
        \end{tabular}\vspace{2mm}
        \caption{\textbf{Regression Loss.} $\ell_1$ loss gains better performance than Smooth $\ell_1$ loss. The disentanglement form further improves detection result.}
        \label{reg_loss}\vspace{-3mm}
    \end{table}

    \begin{table}[t]
    \centering
        \begin{tabular}{c||ccc}
        \hline
        \multirow{2}{*}{Option} & \multicolumn{3}{c}{3D Object Detection / Birds' Eye View}          \\ \cline{2-4} 
                        & Easy                                & Moderate                         & Hard        \\ \hline
        Quaternion      & 13.36 / 17.81                       & 12.52 / 15.16                    & 11.31 / 15.00   \\
        Vectorial     & \textbf{14.76} / \textbf{19.99}     & \textbf{12.85} / \textbf{16.07}  & \textbf{11.50} / \textbf{15.39}   \\ \hline 
        
        \end{tabular} \vspace{2mm}
        \caption{\textbf{Rotation Parametrization.}  Vectorial representation of angles yileds better result than the quaternion representation.}
        \label{rot_param}\vspace{-3mm}
    \end{table}
    In this section, we show the results of experiments we conducted to compare different normalization choices, loss function, and rotation angle parameterizations. All experiments are performed on the \textit{train/val} split on the KITTI dataset. Moreover, we use \textit{car} class to evaluate our model.
    
    \paragraph{Normalization Strategy:} 
    We chose GN as the normalization strategy since it is less sensitive to batch size and cross-GPU training issues. We compare the performance difference in the 3D detection task of BN and GN used in the backbone network. As illustrated in Tab.~\ref{normal_strat}, GN achieves significant improvement over BN on the \emph{val} set. In addition, we notice that GN can save considerable time in training. For each epoch, GN consumes around 5 minutes while BN needs 8 minutes which takes 60\% more time compared to GN.

    \paragraph{Regression Loss:} 
    As shown in Tab.~\ref{reg_loss}, we compare different regression loss functions for 3D bounding box estimation performance. We observe that $\ell_1$ loss performs better than Smooth $\ell_1$ loss. Same phenomenon is also found in the keypoint estimation problem \cite{centernet_2019} where $\ell_1$ loss yields better performance than $\ell_2$ loss. Moreover, applying disentanglement to 3D bounding box regression achieves significantly better performance on both 3D object detection and Birds' eye view evaluation.

    \paragraph{Rotation Parametrization:} 
    We compare the performance of SMOKE with respect to different representations of rotation. Following prior work \cite{roi10d_2019, monodis_2019}, the orientation can be encoded as a 4D quaternion to formulate 3D bounding box. The result with this representation is illustrated in Tab.~\ref{rot_param}. We observe that our simple vectorial representation yields slightly better result than the quaternion representation on both 3D detection and Bird's eye view evaluation.

    \begin{figure*}[t]
        \centering
        \includegraphics[width=0.49\linewidth]{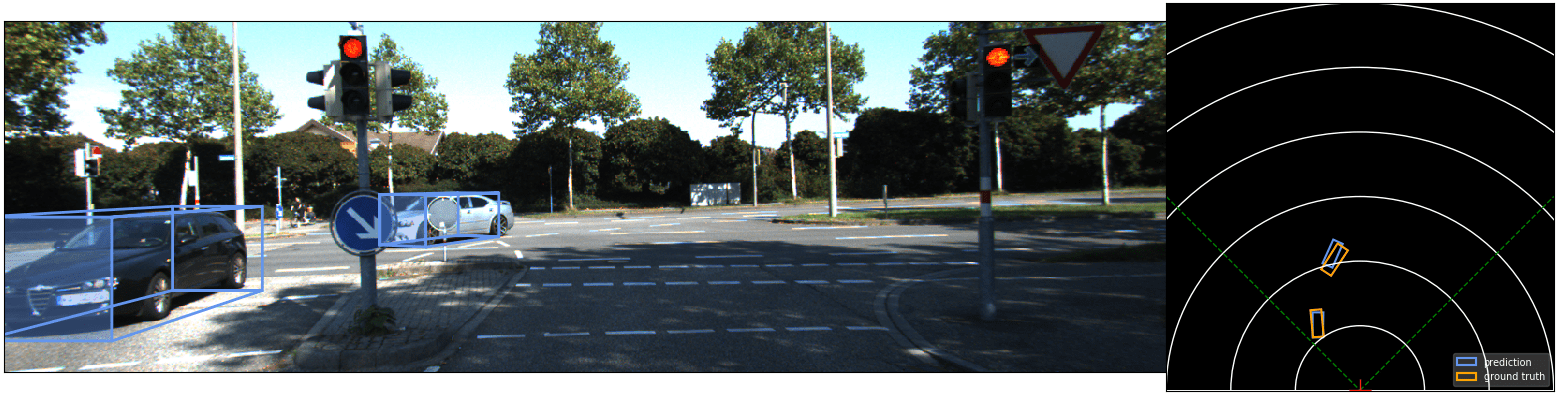}
        \includegraphics[width=0.49\linewidth]{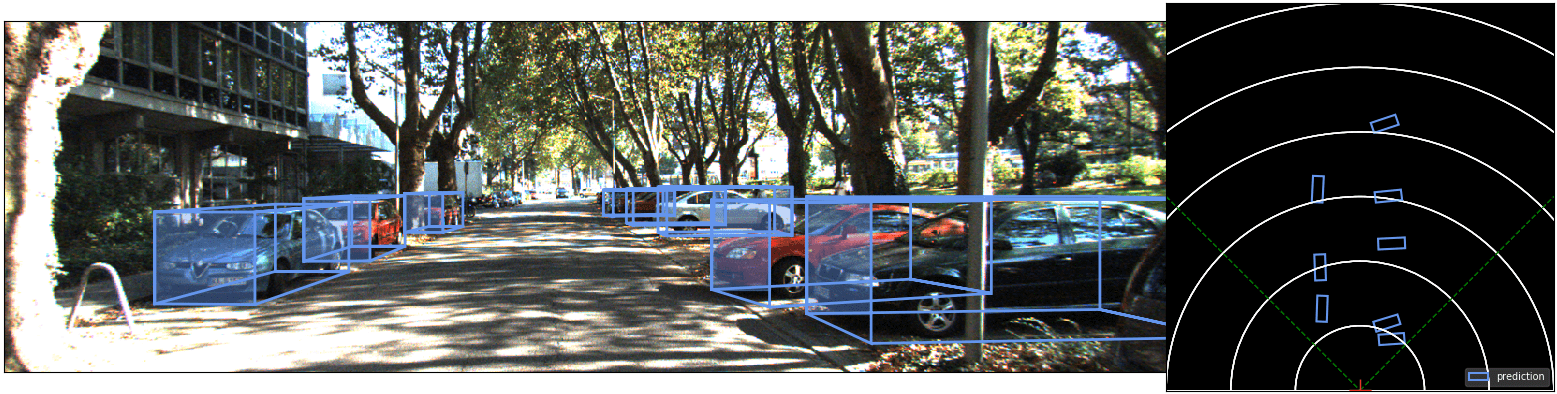}\\ 
        \includegraphics[width=0.49\linewidth]{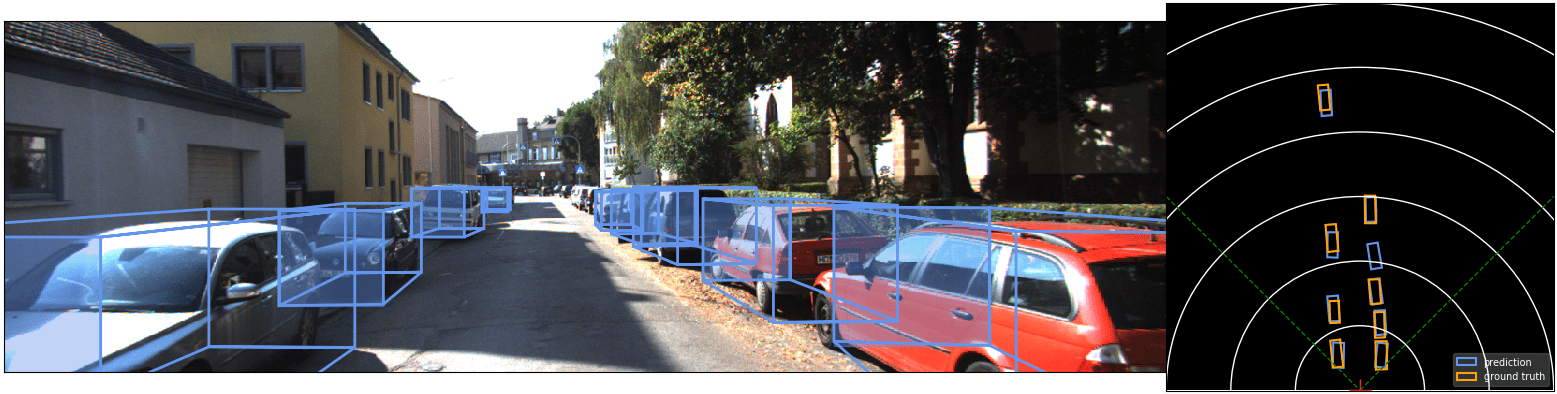}
        \includegraphics[width=0.49\linewidth]{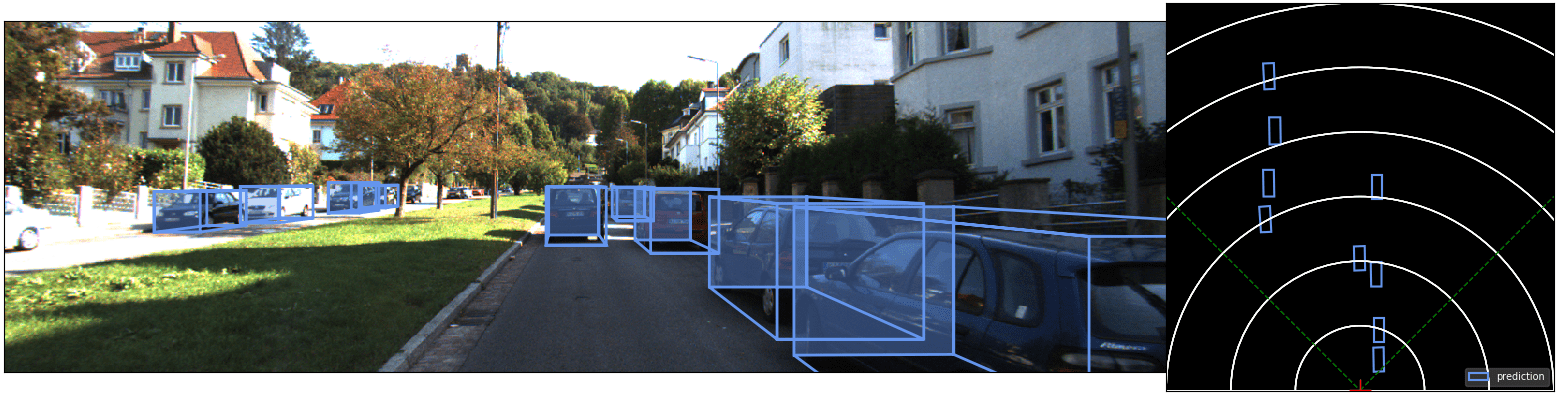}\\
        \includegraphics[width=0.49\linewidth]{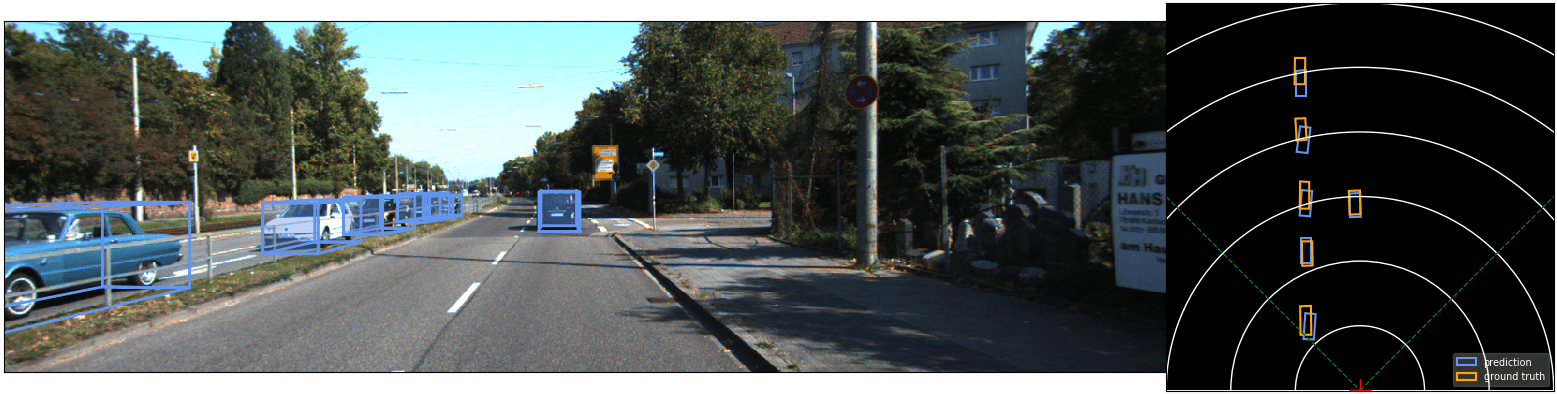}
        \includegraphics[width=0.49\linewidth]{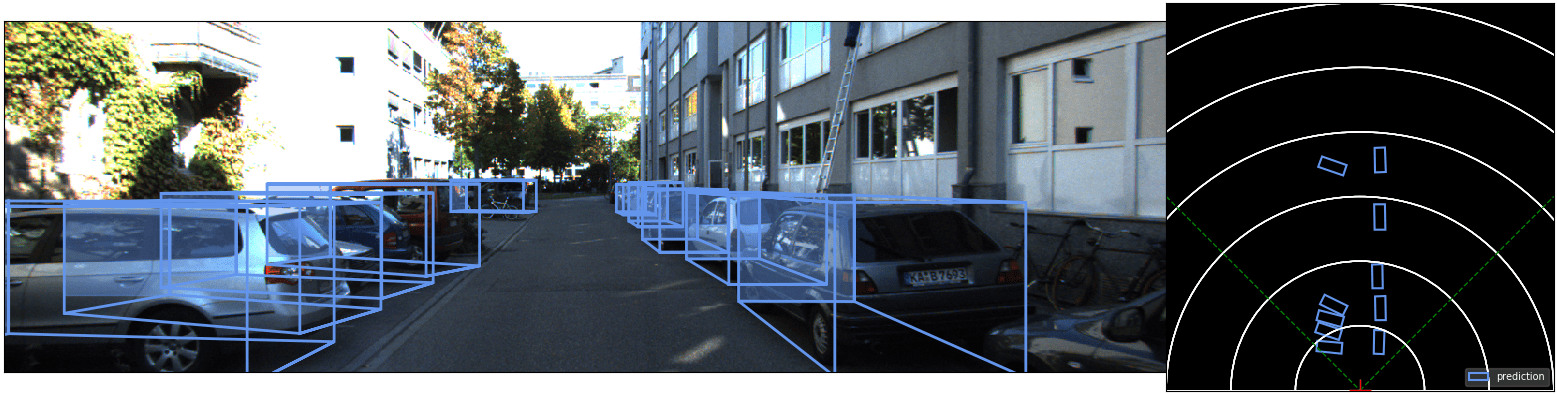}\\ 
        \includegraphics[width=0.49\linewidth]{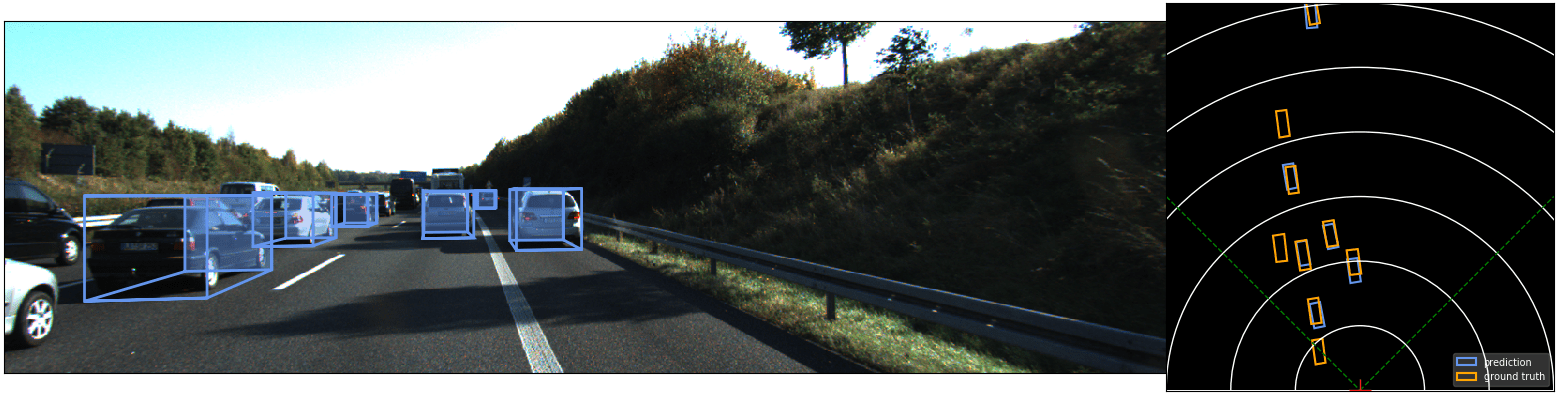}
        \includegraphics[width=0.49\linewidth]{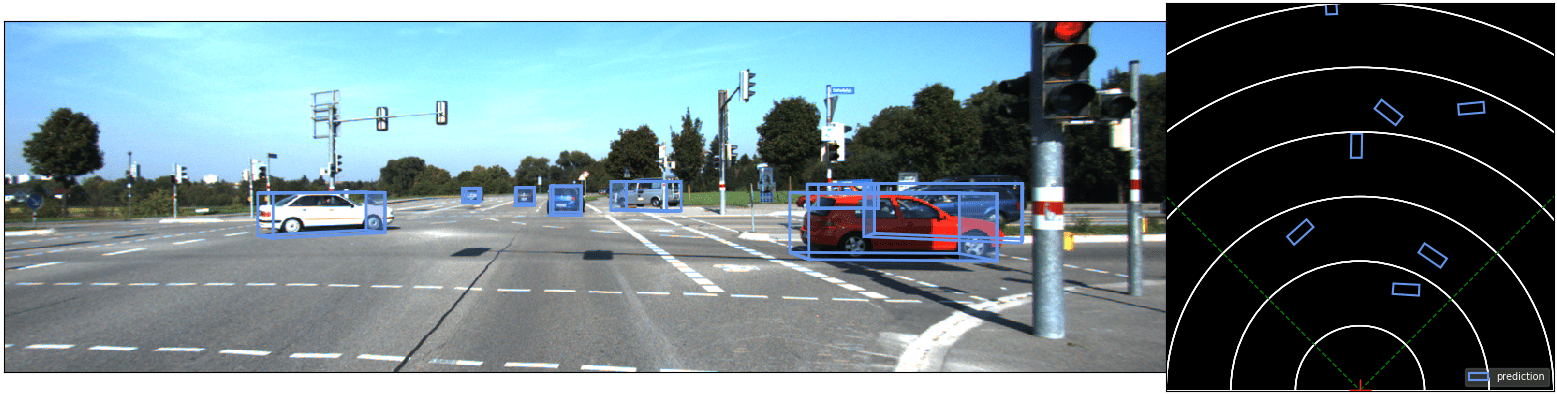}\\
        \includegraphics[width=0.49\linewidth]{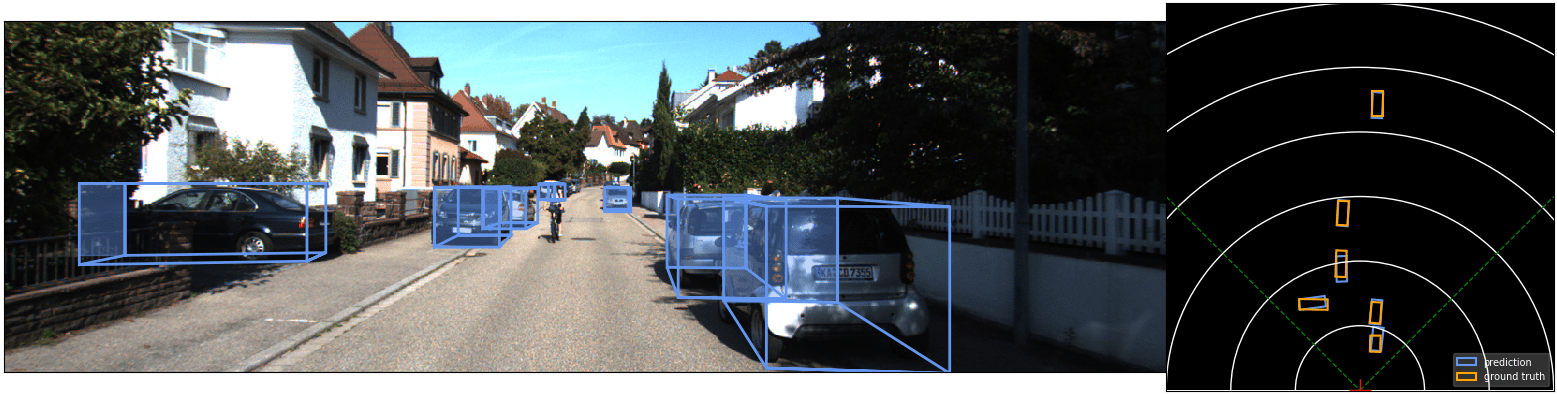}
        \includegraphics[width=0.49\linewidth]{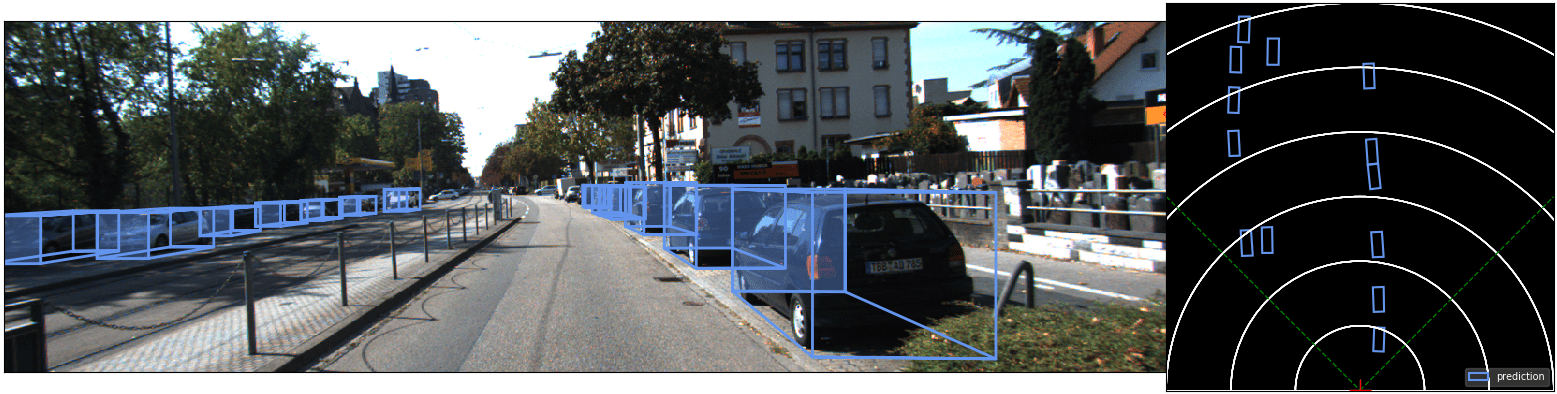}\\
        \includegraphics[width=0.49\linewidth]{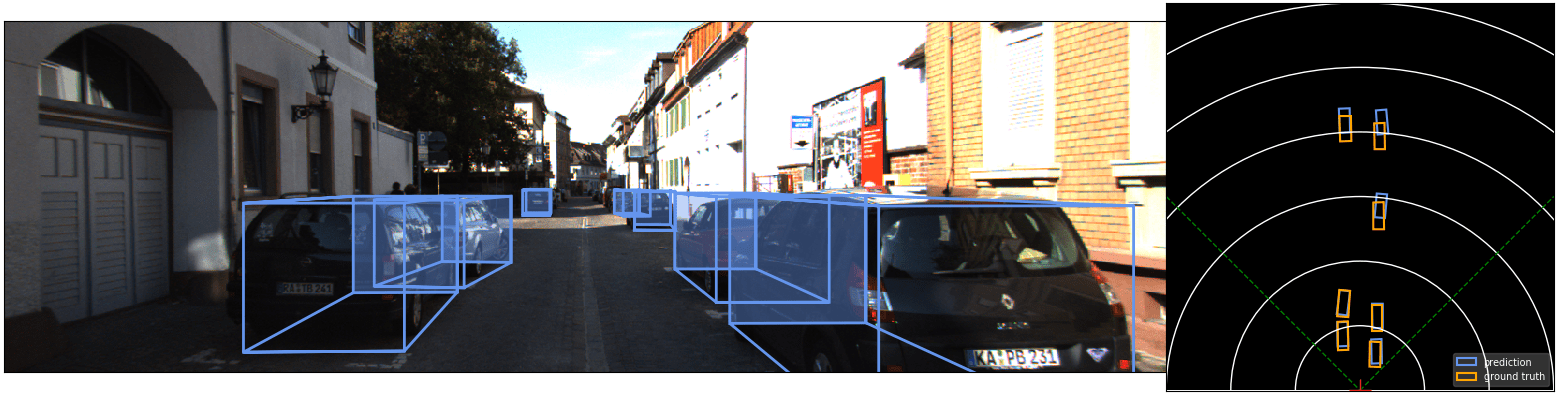}
        \includegraphics[width=0.49\linewidth]{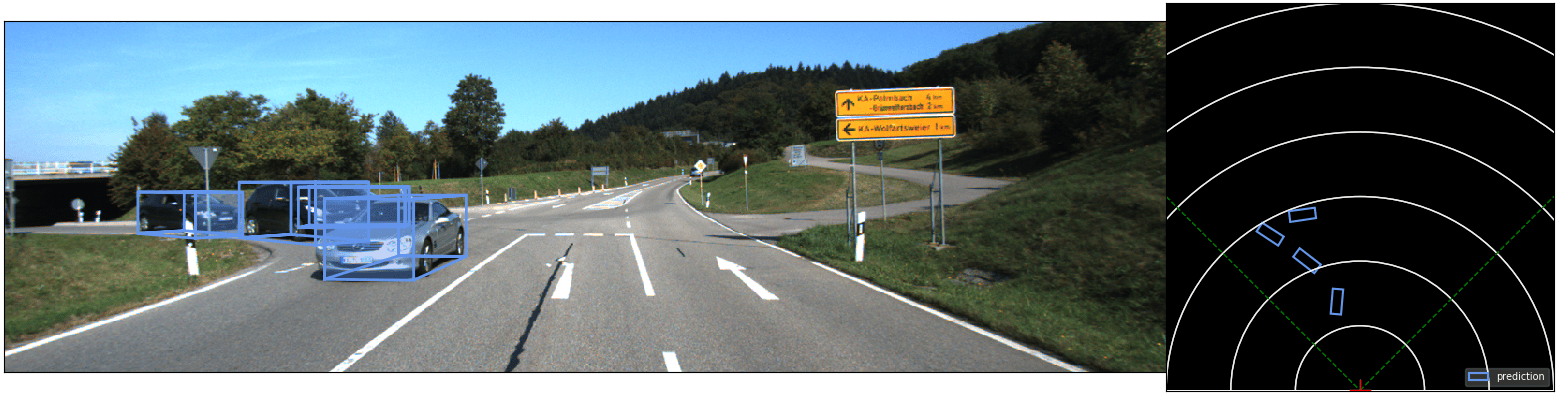}\\
        
        \caption{Qualitative examples from the  validation (left) and test (right) sets in KITTI. The non-transparent side of the bounding box represents the front part of each car. Bird's eye view is also provided to show that SMOKE can recover object distances accurately. Note that all these images are not included in the training phase.}
        \label{quantitative_results}
    \end{figure*}
\subsection{Qualitative Results} 
    
Qualitative results on both the \textit{test} and \textit{val} sets are displayed in Fig.~\ref{quantitative_results}. For better visualization and comparison, we also plot the object localization in Bird's eye view. The results clearly demonstrate that SMOKE can recover object distances accurately

\section{Conclusion and Future Work}
 In this paper, we presented a novel single-stage monocular 3D object detection method based on projected 3D points on the image plane. Unlike previous methods, which depend on 2D proposals to estimate 3D information, our approach regresses 3D bounding boxes directly. This leads to a simple and efficient architecture. To further improve the convergence of regression loss, we proposed a multi-step disentanglement method to isolate the contribution of various parameter groups. In addition, our model does not need synthetic data, complicated pre/post-processing, and multi-stage training. In overall, we largely improve both the detection accuracy and speed on KITTI 3D object detection and Bird's eye view tasks.

Our proposed SMOKE 3D detection framework achieves promising accuracy and efficiency, which can be further extended and used on autonomous vehicles and in robotic navigation. In the future, we aim at extending our method to stereo images and further improving the estimation of projected 3D keypoints and their depth.

{\small
\bibliographystyle{ieee_fullname}
\bibliography{smoke.bib}
}

\end{document}